\documentclass[letterpaper]{article} 
\usepackage{aaai24}  
\usepackage{times}  
\usepackage{helvet}  
\usepackage{courier}  
\usepackage[hyphens]{url}  
\usepackage{graphicx} 
\usepackage{natbib}  
\usepackage{caption} 
\usepackage{algorithm}
\usepackage{algorithmic}

\usepackage{amssymb}
\usepackage{amsmath,amsthm,amsfonts,amssymb,bm}

\usepackage{multirow}
\usepackage{booktabs}

\DeclareMathOperator{\softmax}{SoftMax}

\usepackage{newfloat}
\usepackage{listings}
\DeclareCaptionStyle{ruled}{labelfont=normalfont,labelsep=colon,strut=off} 
\floatstyle{ruled}
\newfloat{listing}{tb}{lst}{}
\floatname{listing}{Listing}
\title{Semantic Lens: Instance-Centric Semantic Alignment for Video Super-Resolution}

\author {
    Qi Tang\textsuperscript{\rm 1,\rm 2},
    Yao Zhao\textsuperscript{\rm 1,\rm 2},
    Meiqin Liu\textsuperscript{\rm 1,\rm 2$ ^*$},
    Jian Jin\textsuperscript{\rm 3},
	Chao Yao\textsuperscript{\rm 4\thanks{Corresponding authors.}}
}
\affiliations {
    \textsuperscript{\rm 1}Institute of Information Science, Beijing Jiaotong University, Beijing, China\\
    \textsuperscript{\rm 2}Beijing Key Laboratory of Advanced Information Science and Network Technology, Beijing, China\\
    \textsuperscript{\rm 3}Alibaba-NTU Singapore Joint Research Institute, Nanyang Technological University, Singapore\\
    \textsuperscript{\rm 4}School of Computer and Communication Engineering, University of Science and Technology Beijing, Beijing, China\\
    \{qitang, yzhao, mqliu\}@bjtu.edu.cn, jian.jin@ntu.edu.sg, yaochao@ustb.edu.cn}

\usepackage{bibentry}

\graphicspath{{figures/}}

\begin{document}

\maketitle

\begin{abstract}
As a critical clue of video super-resolution (VSR), inter-frame alignment significantly impacts overall performance. However, accurate pixel-level alignment is a challenging task due to the intricate motion interweaving in the video. In response to this issue, we introduce a novel paradigm for VSR named \textbf{Semantic Lens}, predicated on semantic priors drawn from degraded videos. Specifically, video is modeled as instances, events, and scenes via a Semantic Extractor. Those semantics assist the Pixel Enhancer in understanding the recovered contents and generating more realistic visual results. The distilled global semantics embody the scene information of each frame, while the instance-specific semantics assemble the spatial-temporal contexts related to each instance. Furthermore, we devise a \textbf{S}emantics-\textbf{P}owered \textbf{A}ttention \textbf{C}ross-\textbf{E}mbedding (SPACE) block to bridge the pixel-level features with semantic knowledge, composed of a \textbf{G}lobal \textbf{P}erspective \textbf{S}hifter (GPS) and an \textbf{I}nstance-Specific \textbf{S}emantic \textbf{E}mbedding \textbf{E}ncoder (ISEE). Concretely, the GPS module generates pairs of affine transformation parameters for pixel-level feature modulation conditioned on global semantics. After that, the ISEE module harnesses the attention mechanism to align the adjacent frames in the instance-centric semantic space. In addition, we incorporate a simple yet effective pre-alignment module to alleviate the difficulty of model training. Extensive experiments demonstrate the superiority of our model over existing state-of-the-art VSR methods.
\end{abstract}

\section{Introduction}

Owing to various constraints, videos captured in real scenes suffer from degradations such as low resolution, blur, and noise, etc. Video super-resolution (VSR) aims to recover high-resolution (HR) video frames from their low-resolution (LR) counterparts with better visual quality. With the proliferation of high-definition displays and the ever-increasing demand for superior video contents, VSR has gained widespread attention. It not only improves the viewer's visual enjoyment, but also holds potential applications in fields like film restoration~\cite{DBLP:conf/cvpr/Wan00022}, medical imaging~\cite{lu2023virtual} and satellite imagery~\cite{deudon2020highresnet}.

Compared with single image super-resolution (SISR), adjacent frames usually contain highly related contents, which can provide complementary information for video super-resolution. Thus, the effective utilization of temporal information across video frames can yield fine-grained visual details. In recent years, an increasing number of VSR approaches can be categorized into two paradigms, explicit and implicit inter-frame alignment. The explicit alignment attempts to align supporting frames with the reference frame by using extracted motion information, including optical flow-based approaches at the image/feature level~\cite{DBLP:conf/cvpr/ChanWYDL21}, optical flow-guided deformable convolution~\cite{DBLP:conf/cvpr/ChanZXL22a}, trajectory-aware attention~\cite{DBLP:conf/cvpr/Liu0FQ22}, etc. The implicit motion estimation and motion compensation seamlessly extract spatio-temporal features from multiple frames like 3D convolution~\cite{DBLP:conf/aaai/LiuZRS021}.

\begin{figure}[!t]
	\centering
	\includegraphics[width=\columnwidth]{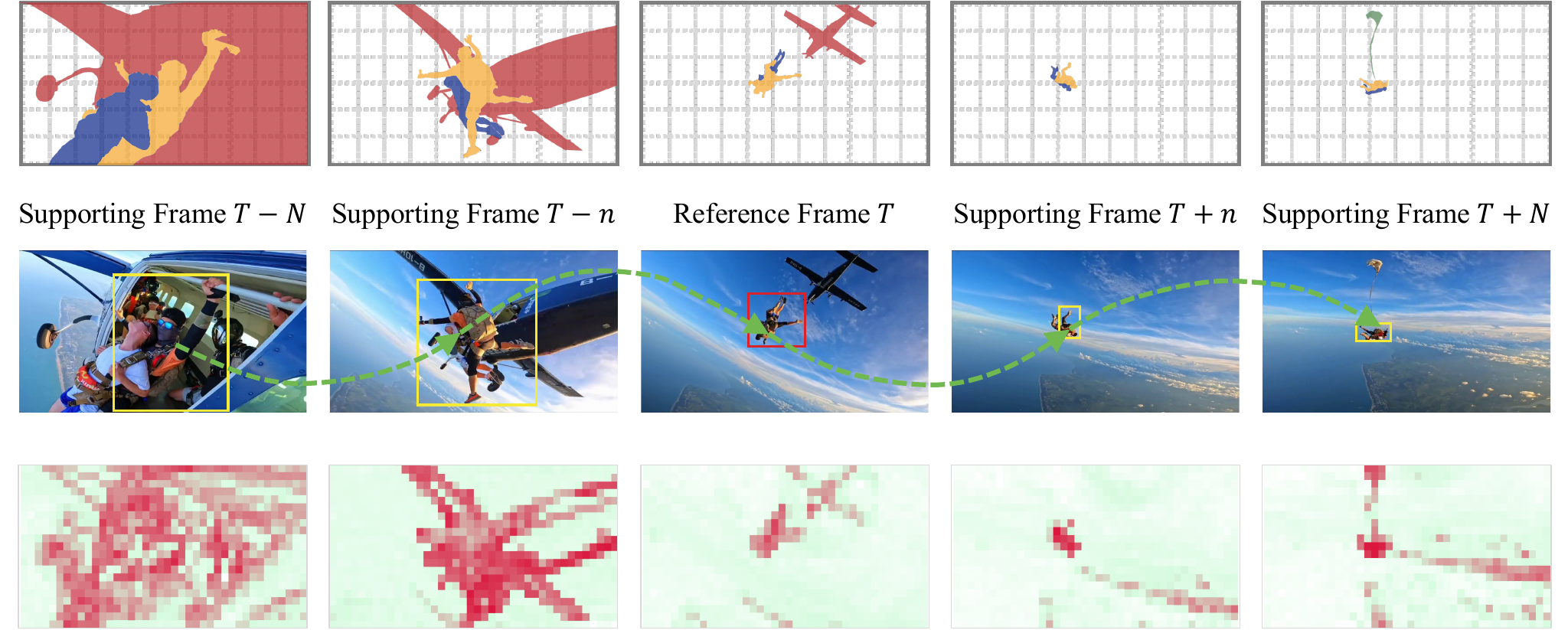}
	\caption{\textbf{Top}: Global semantics and instance-specific semantics. \textbf{Middle}: Original frames and event information (indicated by green arrows). \textbf{Bottom}: Patch PSNR heat map of five frames in a video, super-resolved by a single image super-resolution model. A clear boundary shows that PSNR is strongly related to video content.}
	\label{fig:intro}
\end{figure}

However, in the temporal dimension, accurate pixel-level alignment still poses a challenging task. On one hand, different contents exhibit varying restoration difficulties~\cite{Li_2023_CVPR}. As illustrated in Figure \ref{fig:intro}, smooth regions (indicated in light green) are easy to recover, while textured areas (marked in red) are relatively difficult. The sub-optimal restoration results of texture-rich regions will hinder the accuracy of alignment, and even for easy-to-recover contents, high-frequency artifacts may be introduced due to misalignment~\cite{DBLP:conf/cvpr/ZhouLL0L22}. On the other hand, in many real scenes, the camera's viewpoint is not fixed, leading to significant changes in the background between adjacent frames. Beyond movement from the camera, the scene itself may also be constantly changing with lighting and compositional elements. Moreover, the trajectories of instances within the video are usually inconsistent~\cite{liu2022ttvfi,10255610}. These diverse motion cues, when overlaid, make precise inter-frame alignment difficult. For instance, when facing information loss due to occlusion, pixel-level alignment may bring holes and blur.

To address these issues, we draw inspiration from how humans perceive images, which tend to automatically understand these contents hierarchically. For example, we might intuitively regard the front-running race car on a track as the foreground, while other parts are perceived as the background. Thus, a straightforward approach is to leverage advanced video analysis techniques to model the foreground and background of videos. The intuition behind this is to establish instance-centric semantic representations, and at the same time, to attribute diverse motion information contained in the video to the corresponding instance. Each instance or background is regarded as an independent movement unit. In this way, the original complex intertwined motion is now broken down into several simple and independent movements. Besides, the change of object scale and illumination seriously affects the accuracy of alignment in pixel space, while those factors can be ignored in semantic space. Therefore, the semantic association of same object in temporal dimension can support the alignment of objects in multi-frames, thus avoiding the complex interactions of multi-instance or background noise and the challenge of accurate pixel-level alignment.

Driven by the above analysis, we introduce \textbf{Semantic Lens}, a novel paradigm for video super-resolution. It combines the magnifying power of a lens with the semantic priors, enabling not only improving resolution but also generating enjoyable visual results. Specifically, we decouple the video, originally composed of frames, into instances, events and scenes via constructing a Semantic Extractor. To make use of those semantics, we develop a Semantics-Powered Attention Cross-Embedding (SPACE) block, comprising a Global Perspective Shifter (GPS) and an Instance-Specific Semantic Embedding Encoder (ISEE). It embeds the semantics into features extracted from LR frames in a position-embedding-like manner and enables the instance-centric inter-frame alignment with the guidance of semantic priors. Experimental results demonstrate that our method consistently outperforms existing state-of-the-art methods. The main contributions are summarized as follows:

\begin{itemize}
	\item We pioneer a novel representation for video modeled as instances, events, and scenes, providing both global semantics and instance-specific semantics to boost the performance of video super-resolution.
	\item We propose a Semantics-Powered Attention Cross-Embedding block to bridge semantic priors and pixel-level features, being aware of the restored contents. 
	\item We further design Instance-Specific Semantic Embedding Encoder to perform inter-frame alignment in the instance-centric semantic space via attention mechanism.
\end{itemize}

\section{Related Work}

\noindent\textbf{Video Super-Resolution} Existing VSR framework generally consists of feature extraction, alignment, fusion and frame reconstruction. It is vital for them to utilize the information of inter-frame to improve their performance. For example, EDVR~\cite{DBLP:conf/cvpr/WangCYDL19} adopts multi-scale deformable convolution (DCN) and pyramid structure to enhance the alignment. BasicVSR~\cite{DBLP:conf/cvpr/ChanWYDL21} experimentally verifies that feature-level warping can alleviate the loss of details caused by the inaccurate optical flow estimation. BasicVSR++~\cite{DBLP:conf/cvpr/ChanZXL22a} proposes flow-guided deformable alignment to stabilize the training of DCN, which explores multiple related pixels to reduce artifacts. TCNet~\cite{DBLP:journals/tcsv/LiuJYLZ23} devises a spatio-temporal stability module to enhance the structure stability of frames and the temporal consistency of the video. RVRT~\cite{DBLP:conf/nips/LiangFXRIGC0TG22} divides the video sequence into multiple clips and transmits information clip-by-clip, achieving the balance between performance and computational complexity. Reporting that inaccurate optical flow estimation and resampling operations will destruct the sub-pixel information in the LR video, PSRT~\cite{DBLP:conf/nips/ShiGXWYD22} proposes patch alignment. TTVSR~\cite{DBLP:conf/cvpr/Liu0FQ22} employs optical flow and attention mechanism to locate the most similar frame patch along the same trajectory and makes full use of the information of the whole sequence. Yet, the performance of optical flow-based methods always degrades greatly to process videos with large motion or significant lighting changes. 

\noindent\textbf{Visual Understanding} Visual understanding at different granularity levels has been a longstanding problem~\cite{zhu2023ctp} in the computer vision field. Segmentation, whose essence lies in grouping pixels, encompasses various tasks based on different semantics. Semantic segmentation seeks to discern semantic categories per-pixel within an image~\cite{zhang2022mining}, while instance segmentation~\cite{DBLP:conf/iccv/HeGDG17} clusters pixels with identical semantic connotation into objects. With the ability to understand at both the pixel and instance levels, it is logical to step toward the panoptic segmentation~\cite{DBLP:conf/cvpr/KirillovHGRD19}. Lately, a tendency has emerged to design a universal architecture capable of handling all segmentation tasks, like Mask2Former~\cite{DBLP:conf/cvpr/ChengMSKG22}, which excels over specialized architectures in various segmentation tasks while being straightforward to train.

 \begin{figure*}[!htbp]
	\centering
	\includegraphics[width=0.96\linewidth]{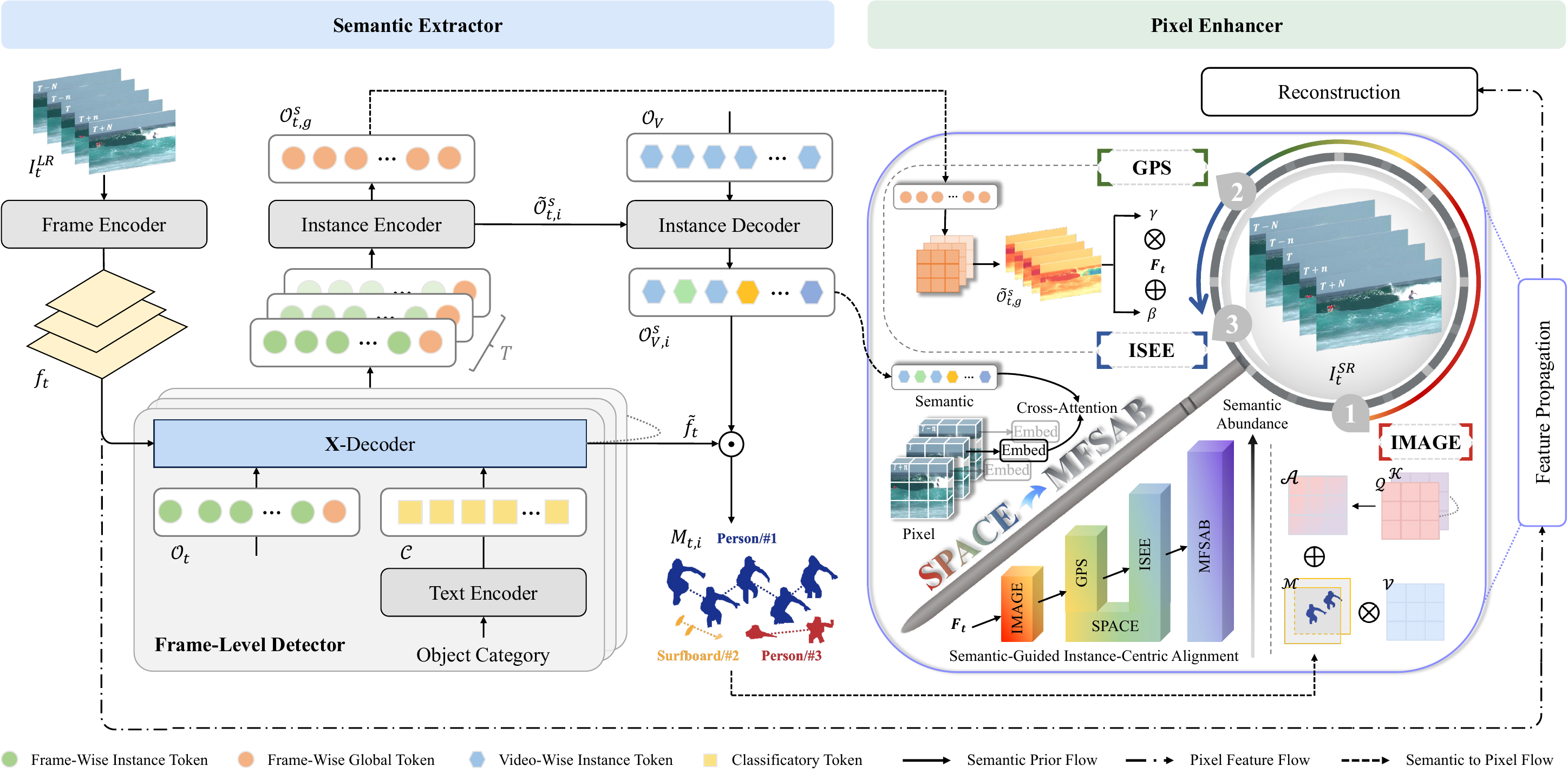}
	\caption{Overall pipeline for \textbf{Semantic Lens} consists of a Semantic Extractor and a Pixel Enhancer. The Semantic Extractor decouples low-resolution video into instances, events, and scenes, each characterized by their embodied semantics with differentiated descriptors. These semantics are employed to enhance the pixel-level features of Pixel Enhancer in a position-embedding-like manner, which yields semantic-aware features.}
	\label{fig:model}
\end{figure*}

In this paper, we forgo the pixel-level inter-frame alignment dominated by optical flow, favoring a content-oriented alignment based on semantic cues instead. Our method outshines state-of-the-art works in several evaluation metrics.

\section{Method}

The proposed paradigm for video super-resolution, named \textbf{Semantic Lens}, is illustrated in Figure \ref{fig:model}, which consists of a Semantic Extractor and a Pixel Enhancer. Let $I^{LR}=\left\{I^{LR}_t|t\in\left[1,T\right]\right\}$ be a LR video sequence of height $H$, width $W$, and frame length $T$. The Pixel Enhancer aims to recover a high-resolution version, i.e., $I^{HR}=\left\{I^{HR}_t|t\in\left[1,T\right]\right\}$ of height $sH$, width $sW$, in which $s$ is the up-sampling factor. It coincides with the mission of the conventional VSR network. Differently, semantic priors derived from Semantic Extractor are embedded into the features of the Pixel Enhancer for instance-centric alignment, generating more enjoyable visual results. In this section, we will elaborate on the detailed designs of diverse semantics' utilization as well as the overall architecture.

\subsection{Architecture}

At each time step $t$, given a reference frame $I_t^{LR}$ and $2n$ adjacent supporting frames $\{I_{t-n}^{LR},\cdots,I_{t-1}^{LR},I_{t+1}^{LR},\cdots,I_{t+n}^{LR}\}$, the Semantic Lens delivers the super-resolved reference frame $I_t^{SR}$ as output. First, the Semantic Extractor is responsible to extract diverse semantic priors from LR frames, which is mainly composed of two parts, that is, frame encoder-decoder and instance encoder-decoder. The frame encoder-decoder spatially models intra-frame correlations frame by frame, obtaining global information and instance-specific context at the frame level. While the instance encoder-decoder allows the interaction of the instance-related content along the time dimension, and establishes the video-wise instance representation under the supervision of video instance segmentation (VIS). In this way, the video content is encoded into the instance-centric semantic space. And then, those semantic information serves as the input of the subsequent stage together with the LR frames.

Pixel Enhancer is in charge of refining the original pixels and generating absent information. It adopts the bidirectional and second-order grid propagation identical to that of BasicVSR++, thereby enabling aggressive exploitation of information from the entire video. The bidirectional feature propagation branch is composed of Multi-Frame Self-Attention Blocks (MFSAB)~\cite{DBLP:conf/nips/ShiGXWYD22}. Walking past the stacked MFSABs, the refined deep features $\boldsymbol{F_t}$ of the reference frame are used to generate the high-resolution result via a reconstruction module, consisting of several convolution layers as well as pixel-shuffle layers. 

With the aim of improving the inter-frame alignment, we formulate a semantic-guided instance-centric alignment schema. Specifically, features of supporting frames are first coarsely warped to reference frame via the IMAGE module. Then, the designed SPACE block is positioned before the basic unit of the feature propagation branch, and iteratively enhances the semantic abundance of original features. Based on global semantics, the GPS module generates pairs of scale and bias for spatially-adaptive feature modulation. The modulated features are further aligned with the guidance of instance-specific semantics by the ISEE module, which means that the inter-frame alignment is constrained within the same instance and its related context. Finally, the features aligned in instance-centric semantic space are fed into the subsequent MFSAB. %

\subsection{Semantic Extractor}

\begin{figure*}[!t]
	\centering
	\includegraphics[width=0.96\linewidth]{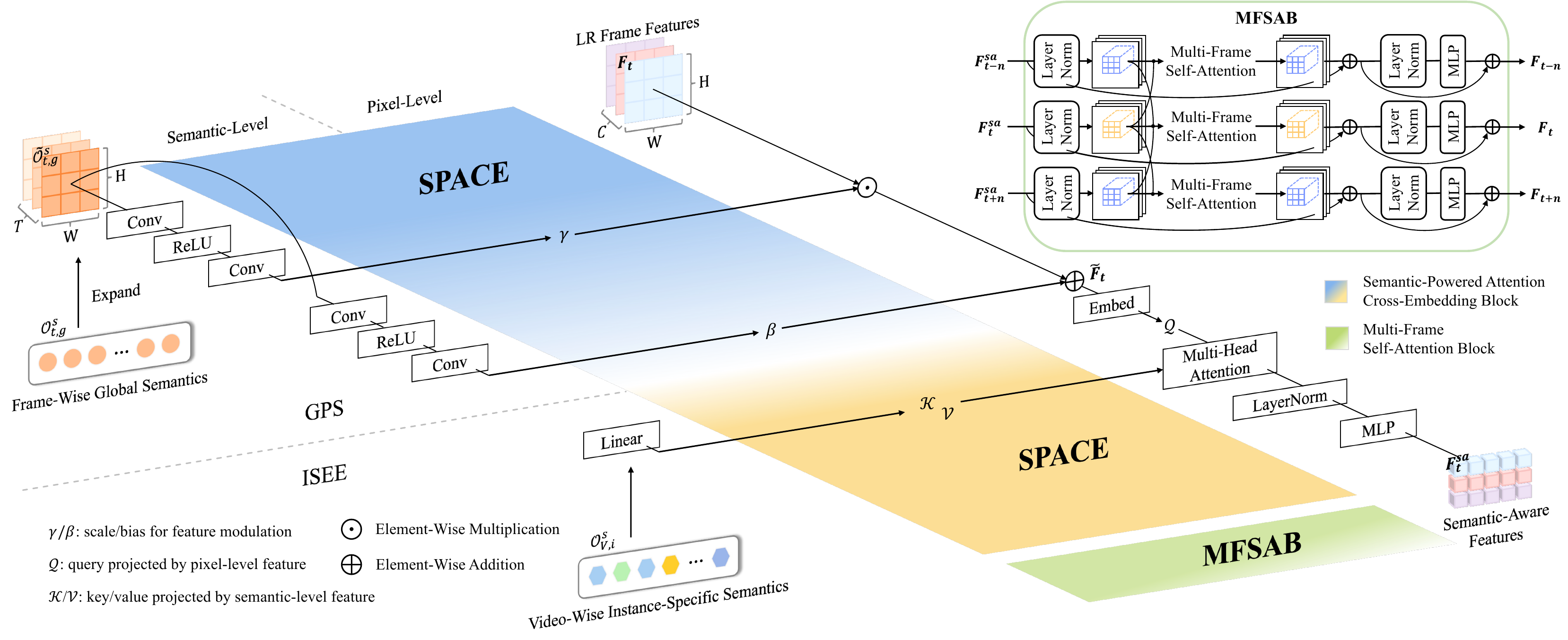}
	\caption{Illustration of Semantics-Powered Attention Cross-Embedding (SPACE) Block, composed of Global Perspective Shifter (GPS) and Instance-Specific Semantic Embedding Encoder (ISEE). It is inserted before MFSAB, the basic unit of feature propagation in Pixel Enhancer, to bridge the semantic-level priors with pixel-level features.}
	\label{fig:space}
\end{figure*}

The Semantic Extractor aims to encode the pixel-based representation of a video into a latent semantic space. In this semantic space, video is decoupled into instances (foreground), events (temporal), and scenes (background). Among them, instances and scenes symbolize the characters and locations recorded in the video, while events encompass the context of instances and their interactions across frames. To put it another way, the video is transformed into the instance-centric form. In this work, we employ X-Decoder~\cite{Zou_2023_CVPR}, a frame-wise detector, to distill instance and scene information of each frame into tokens, which can be formulated as:

\vspace{-0.5cm}
\begin{small}
	\begin{gather}
	\small
	f_t =\mathbf{Enc}_{frame}(I^{LR}_t)\\
	\mathcal{O}_t^s,\tilde{f}_t=\mathbf{XDec}(f_t,\mathcal{O}_t,\mathcal{C})		
\end{gather}
\end{small}

\vspace{-0.15cm}
\noindent where $\mathbf{Enc}_{frame}$ and $\mathbf{XDec}$ separately denote the frame encoder and decoder. $f_t$ and $\tilde{f}_t$ are the pixel-level features generated by the encoder and decoder respectively. The generic non-semantic queries $\mathcal{O}_t$ and classificatory queries $\mathcal{C}$ serve as the inputs of the decoder together with $f_t$. $\mathcal{O}_t$ is randomly initialized, while $\mathcal{C}$ is obtained by encoding the category labels through a text encoder. $\mathcal{O}_t^s$ represents the token-level semantics, in which the last semantic token gleans the global image representation, with the rest being used for instance segmentation. 

For the event, we gather instance-specific tokens from all frames to construct the video-wise instance representation. It can be used for generation of consistent instance masks given specific frame in VIS, thus it can also locate the pixels related to the same instance of multiple frames for alignment in VSR. Specifically, instance-specific tokens are divided into different local temporal windows without overlapping. In the instance encoder, window-based self-attention is employed upon instance-specific tokens which shifts along the temporal dimension, so that tokens from different frames can communicate instance-specific information. The above process can be formulated as:

\vspace{-0.35cm}
\begin{small}
\begin{equation}
	\mathcal{\tilde{O}}_{t,i}^s = \mathbf{Enc}_{inst}(\mathcal{O}_{t,i}^s), i\in \left[1,N_f\right]
\end{equation}
\end{small}

\vspace{-0.05cm}
\noindent where $\mathbf{Enc}_{inst}$ denotes the instance encoder. $\mathcal{O}_{t,i}^s, \mathcal{\tilde{O}}_{t,i}^s$ represent the $i$-th instance-specific token within $t$-th frame and those have interacted with each other along the temporal axis in instance encoder, respectively. $N_f$  is the number of instance-specific tokens in each frame.

The construction of video-wise instance representation is to incorporate the appearance of instances across multiple frames, i.e., a more essential depiction of instance. For this reason, trainable video queries $\mathcal{O}_V$ are used to embody the video-wise instance-specific semantics and event semantics:

\vspace{-0.4cm}
\begin{small}
\begin{gather}
	\mathcal{O}_{V,i}^s = \mathbf{Dec}_{inst}(\mathcal{\tilde{O}}_{t,i}^s,\mathcal{O}_V)\\
	M_{t,i} = \mathcal{O}_{V,i}^s \cdot \tilde{f}_t
\end{gather}
\end{small}

\vspace{-0.15cm}
\noindent where $\mathbf{Dec}_{inst}$ denotes the instance decoder and $\mathcal{O}_{V,i}^s$ are the video-wise instance-specific representations. Both instance encoder and decoder consist of a stack of blocks based on the attention mechanism. $M_{t,i}$ represents the segmented masks of instances. With an eye to the limited expressiveness of the global semantic token and the intricate scenes of long video sequences, we suggest retaining the global representations of all frames to describe the scene information of the video, denoted as $\mathcal{O}_{t,g}^s$.

\subsection{Semantics-Powered Attention Cross-Embedding}

When aligning inter-frame features with the guidance of semantic priors, a challenge that should be considered is the discrepancy between them. To alleviate this issue, we propose the Semantics-Powered Attention Cross-Embedding (SPACE) Block, as graphically depicted in Figure \ref{fig:space}. The SPACE block is like the bridge between Semantic Extractor and Pixel Enhancer, embedding semantic priors into pixel-level features in a position-embedding-like manner. It is composed of the Global Perspective Shifter (GPS) and the Instance-Specific Semantic Embedding Encoder (ISEE). The former generates pairs of affine transformation parameters for spatial-wise feature modulation conditioned on global semantics, while the latter utilizes the attention mechanism to enable the model to be aware of the specific content during alignment, yielding richer semantic-aware features. 

Specifically, the global semantics of each frame is condensed into a single token, providing the most concise semantic representation of a video frame. GPS firstly extends the global semantics to match the spatial dimensions of the features from the last MFSAB using a simple dot product:

\vspace{-0.25cm}
\begin{small}
\begin{equation}
	\tilde{\mathcal{O}}^{s}_{t,g}=\mathcal{O}_{t,g}^{s}\cdot \boldsymbol{F_t}
\end{equation}
\end{small}

\vspace{-0.05cm}
\noindent where $\mathcal{O}^{s}_{t,g} \in \mathbb{R}^{1\times C}$ and $\tilde{\mathcal{O}}^{s}_{t,g} \in \mathbb{R}^{C\times H \times W}$ are original and extended global semantics of $t$-th frame, respectively. $C$ is the embedding dimension of features during propagation.

\begin{figure}[!tbp]
	\centering
	\includegraphics[width=0.9\columnwidth]{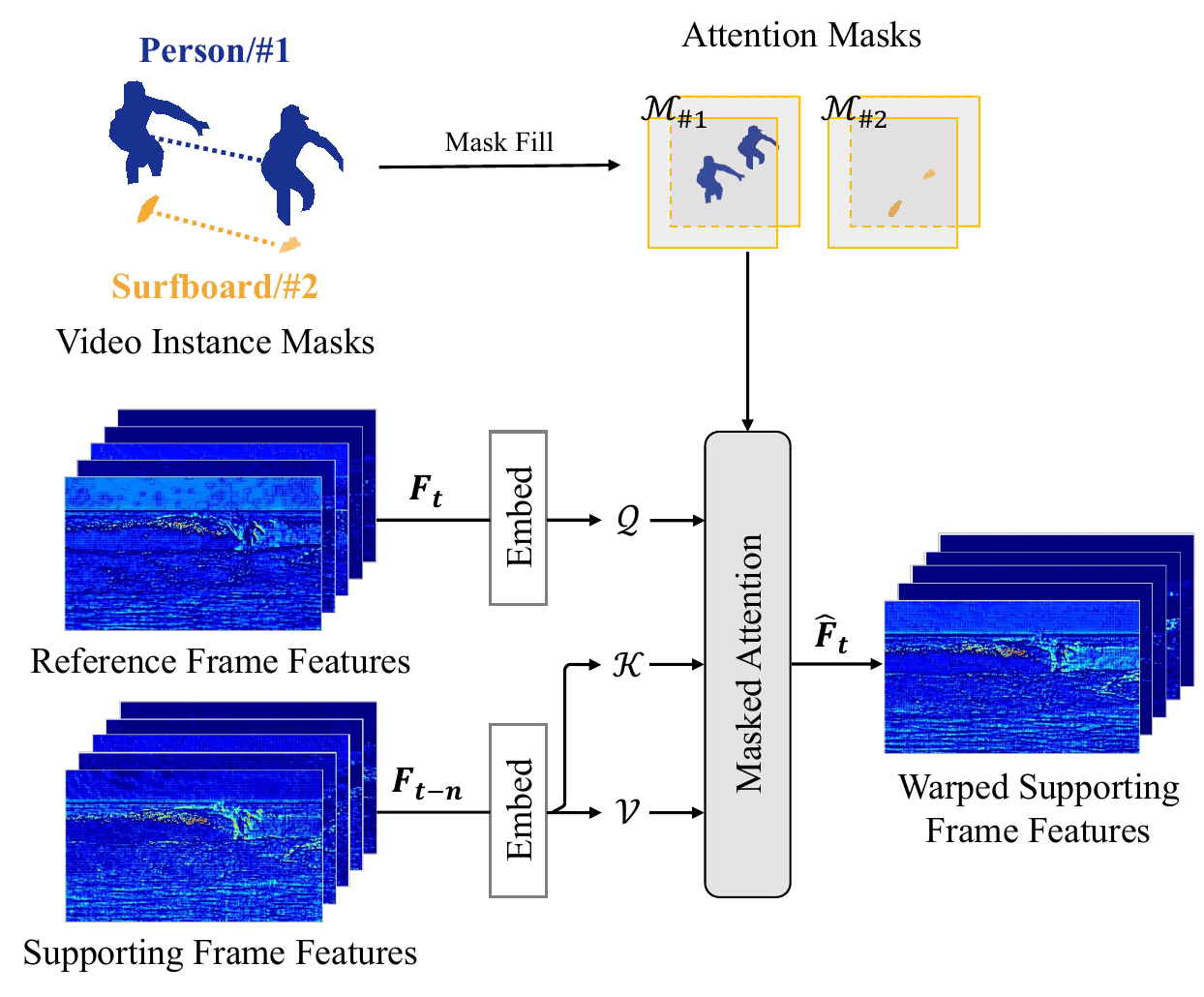}
	\caption{Illustration of Implicit Masked Attention Guided Pre-Alignment (IMAGE) Module. Attention is conducted within a local window for pre-alignment, which is modulated by the instance masks.}
	\label{fig:image}
\end{figure}

Utilizing two convolutional layers followed by a ReLU activation, it generates parameter pairs $(\gamma, \beta)$, which are used for spatial affine transformations of the pixel-level features, representing the scale and bias, respectively:


\vspace{-0.4cm}
\begin{small}
\begin{gather}
		(\gamma,\beta) = \mathcal{G}(\tilde{\mathcal{O}}^{s}_{t,g})\\
		\boldsymbol{\tilde{F}_t} = (\boldsymbol{F_t}\odot \gamma+\beta) +\boldsymbol{F_t}
\end{gather}
\end{small}

\vspace{-0.15cm}
\noindent where $\mathcal{G}$ represents the convolution layers and ReLU activation. $\boldsymbol{\tilde{F}_t}$ denotes the features embedded with global semantics, which are then fed into the ISEE. The intuition behind this methodology lies in allowing each pixel to be modulated by the global semantic prior by generating different modulation parameters, enabling adaptive feature enhancement.

ISEE is built upon the cross-attention, where the queries $\mathbf{Q}$ are derived from pixel-level features, while the keys $\mathbf{K}$ and values $\mathbf{V}$ are generated by instance-specific semantics. It bridges pixel-level features with semantic-level priors by learning the interrelationships between different pixels and instances, thereby embedding instance-specific priors into the original features via attention map $\mathbf{A}$, which is applied to values for weighted summation. The above procedures are formulated as:

\vspace{-0.45cm}
\begin{small}
\begin{gather}
	\mathbf{Q}=\boldsymbol{\tilde{F}_t}W^Q, \mathbf{K}=\mathcal{O}^{s}_{V,i}W^K, \mathbf{V}=\mathcal{O}^{s}_{V,i}W^V\\
		\mathbf{A}=\softmax(\mathbf{Q}\mathbf{K}^\mathrm{T}/\sqrt{D})\\
			\boldsymbol{F_t^{sa}} = \mathbf{A}\mathbf{V}
\end{gather}
\end{small}

\vspace{-0.15cm}
\noindent where $W^Q, W^K, W^V \in \mathbb{R}^{C\times D}$ represent the parameter matrices of projections, and $D$ is the channel number of projected vectors. The refined semantic-aware features $\boldsymbol{F_t^{sa}}$ will serve as the input of the subsequent MFSAB.

Aside from the SPACE, we further introduce a pre-alignment module, named \underline{I}mplicit \underline{M}asked \underline{A}ttention \underline{G}uided Pr\underline{e}-Alignment (IMAGE), as depicted in Figure \ref{fig:image}. IMAGE utilizes the pixels from the reference frame as queries and performs a local window search in the corresponding positions of the supporting frames. By applying attention operations, the supporting frames are coarsely aligned to the reference frame, unlike previous works~\cite{cao2022reference, yang2020learning} which use attention to directly enhance low-resolution images with high-resolution textures. Drawing inspiration from DCN v2, we follow the idea of modulation mask to limit the magnitude of offsets, preventing the model from learning overly irrelevant content. As previously stated, the Semantic Extractor outputs both pixel-level masks and semantic-level tokens simultaneously. Instance masks, which explicitly describe the content correspondence across different frames, naturally serve as modulation masks for IMAGE and are thus used as attention masks for frame-wise pre-alignment. The benefits are two-fold. Firstly, when the pre-aligned supporting frame conducts window-based attention with the reference frame in MFSAB, it provides a broader range of information. Secondly, using instance masks as constraints can avoid performance deterioration caused by irrelevant content, meaning that coarse alignment is executed only between the same instances or backgrounds across frames. Our masked attention modulates the attention matrix via

\vspace{-0.25cm}
\begin{small}
\begin{equation}
	\mathbf{A}=\softmax\left(\mathcal{M}+\mathbf{Q} \mathbf{K}^{\mathrm{T}}\right)
\end{equation}
\end{small}

\vspace{-0.08cm}
\noindent where the attention mask $\mathcal{M}$ at feature location $(x, y)$ is

\vspace{-0.4cm}
\begin{small}
\begin{equation}
	\mathcal{M}(x, y)=\left\{
	\setlength{\arraycolsep}{2.5pt}
	\begin{array}{ll}
0 & \text { if } M_{t,i}(x, y) \bigcap M_{t',i}(x,y) \neq \emptyset \\[0.3em]
-\infty & \text { otherwise }
\end{array}
\right.
\end{equation}
\end{small}

\vspace{-0.15cm}
Here, $M_{t,i}$ and $M_{t',i}$ indicate the $i$-th instance masks from the reference frame and supporting frame, respectively.

\section{Experiments}
\begin{table*}[!t]
  \small
  \centering
  \setlength{\tabcolsep}{1.4mm}{
  \begin{tabular}{l||c||ccc||ccc}
     \hline
    \multirow{2}{*}{Method} & \multirow{2}{*}{Frames} & \multicolumn{3}{c||}{BI degration} & \multicolumn{3}{c}{BD degradation} \\
     & & YTVIS-19 & YTVIS-21 & YTVIS-22  &  YTVIS-19 & YTVIS-21 & YTVIS-22 \\
    \hline
    Bicubic             & - &   32.41/0.9103 & 31.96/0.8874  &  31.92/0.8834  &    31.89/0.8645 & 31.42/0.8567  &  31.33/0.8524  \\   
     \hline
     EDVR~\cite{DBLP:conf/cvpr/WangCYDL19} & 5 & 35.77/0.9429 & 35.26/0.9394  &  35.15/0.9358 & 32.28/0.8667 &    31.90/0.8598 & 31.81/0.8599   \\
     BasicVSR~\cite{DBLP:conf/cvpr/ChanWYDL21}  & 15 &  35.69/0.9446 & 35.35/0.9417  &  34.56/0.9369 & 32.25/0.8630 &    31.90/0.8562 & 31.81/0.8523  \\
     IconVSR~\cite{DBLP:conf/cvpr/ChanWYDL21}    & 15 &   36.25/0.9488 & 35.80/0.9457  &  35.69/0.9424 & 32.21/0.8625   &  31.83/0.8560 & 31.73/0.8519  \\
     BasicVSR++~\cite{DBLP:conf/cvpr/ChanZXL22a}   & 30 &  36.33/0.9493 & 35.89/0.9463 &  35.77/0.9430 & 32.26/0.8645   &  31.88/0.8579 & 31.78/0.8539  \\
     \hline
     TTVSR~\cite{DBLP:conf/cvpr/Liu0FQ22}  & 50 &   36.33/0.9492 & 35.91/0.9465  &  35.81/0.9433 & 32.30/0.8645 &    31.93/0.8578  & 31.82/0.8534 \\
     RVRT~\cite{DBLP:conf/nips/LiangFXRIGC0TG22}  & 30 &  36.53/0.9507 & 36.06/0.9481  &  35.93/0.9444 & 32.31/0.8650  &  31.95/0.8584 & 31.85/0.8544 \\
     PSRT~\cite{DBLP:conf/nips/ShiGXWYD22}  & 16 &  \underline{36.58}/\underline{0.9514} & 36.14/0.9487  &  36.04/0.9458  & \underline{34.95}/\underline{0.9103} &  \underline{32.07}/\underline{0.8619} & \underline{31.98}/0.8582 \\
     IART~\cite{xu2023implicit}  & 16 &   36.55/0.9513 & \underline{36.24}/\underline{0.9491}   &    \underline{36.12}/\underline{0.9460} & 32.45/0.8687  &  32.05/\underline{0.8619} & 31.97/\underline{0.8585} \\
    Semantic Lens (Ours) & 5 & \textbf{37.19}/\textbf{0.9542} &    \textbf{36.55}/\textbf{0.9499}  & \textbf{36.39}/\textbf{0.9468} & \textbf{38.49}/\textbf{0.9625}  &  \textbf{37.93}/\textbf{0.9597} & \textbf{37.73}/\textbf{0.9558}  \\
    \hline
  \end{tabular}}
   \caption{Quantitative comparison (PSNR$\uparrow$ and SSIM$\uparrow$) on the YTVIS (2019/2021/2022) dataset for $4\times$ VSR task. The best performance is displayed in \textbf{bold}, and the second best performance is marked in \underline{underline}.}
  \label{tab:experiment}
\end{table*}

\begin{figure*}[!t]
    \centering
        \includegraphics[width=0.95\textwidth]{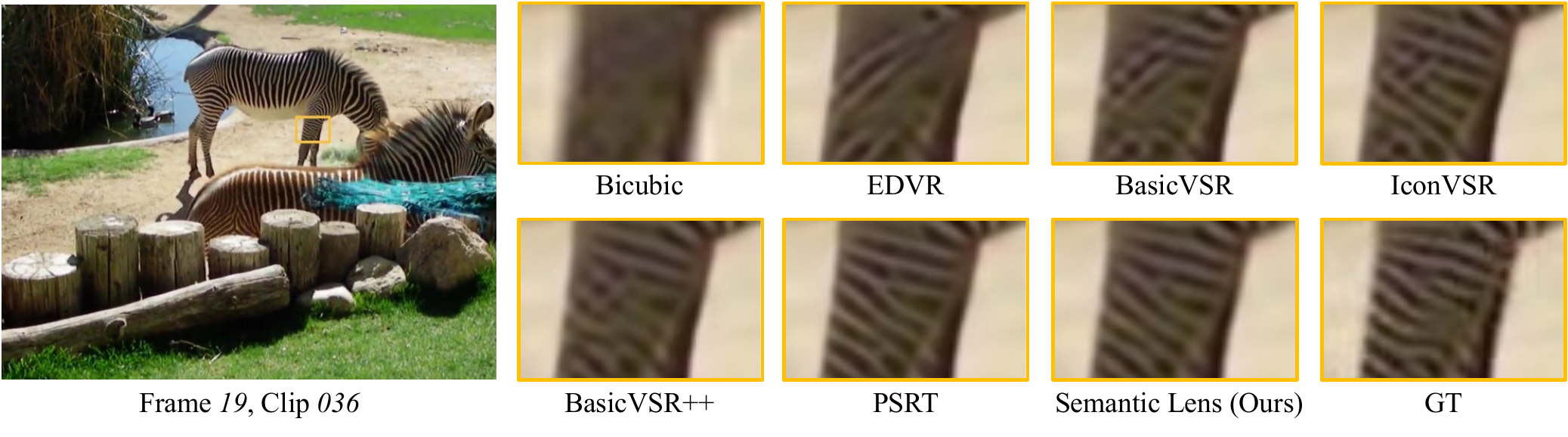}
        \caption{Visual comparison of VSR ($4 \times$) on YTVIS-19 dataset.}
        \label{fig:ytvis}
\end{figure*}

\begin{figure*}[!t]
	\centering
        \includegraphics[width=0.95\textwidth]{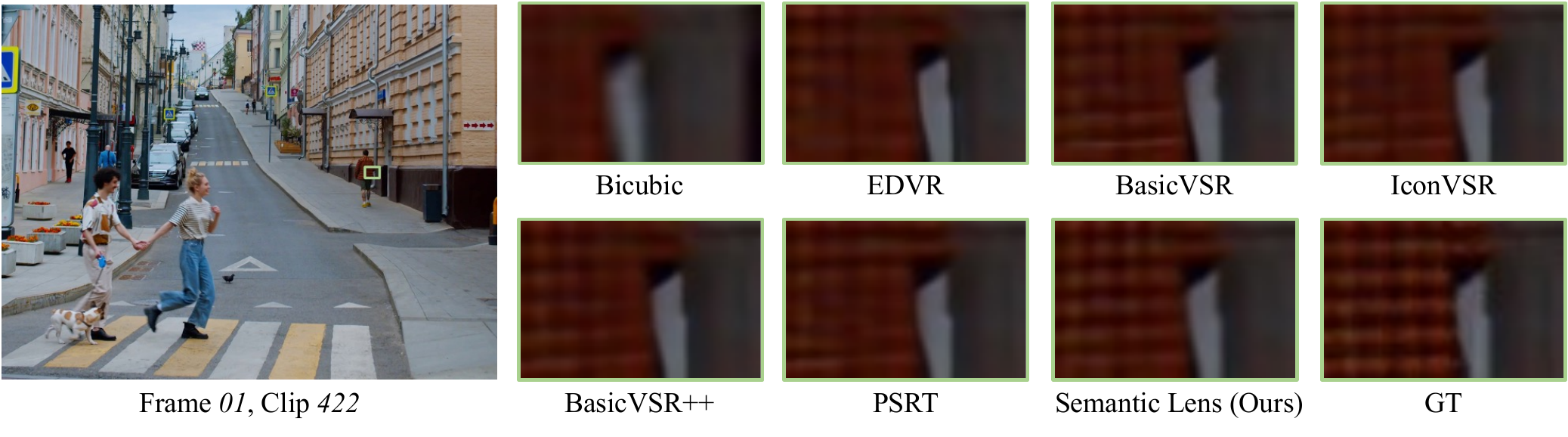}
        \caption{Visual comparison of VSR ($4 \times$) on YTVIS-22 dataset.}
        \label{fig:ytvis-22}
\end{figure*}

\textbf{Datasets and Metrics} We evaluate the performance of Semantic Lens on the benchmark datasets widely used in the field of video instance segmentation, YouTube-VIS (YTVIS)~\cite{DBLP:conf/iccv/YangFX19}, available in three versions (2019, 2021~\cite{ytvis21dataset} and 2022~\cite{ytvis22dataset}). They are chosen for our research with the intent of aiding semantic priors for video super-resolution. YTVIS encapsulates 40 different predefined object categories. YTVIS-19 consists of 2,238 high-resolution video clips for training and 302 for validation, with a total of 4,883 unique video instances. The improved and extended successors, YTVIS-21 and YTVIS-22, share a training set that comprises 2,985 videos. Moreover, additional videos are included in YTVIS-21 for validation, nearly doubling the annotation quantity compared to its 2019 predecessor. On the basis of YTVIS-21, YTVIS-22 introduces 71 more complex and longer videos into the validation set. In consistency with prior studies, we evaluate our model with $4\times$ downsampling using two degradations: Bicubic downsample (BI) and downsampling employing a Gaussian filter with a standard deviation of $\sigma = 1.6$ (BD). We keep the same evaluation metrics: 1) peak signal-to-noise ratio (PSNR) and 2) structural similarity index (SSIM) as previous works.

\noindent\textbf{Model Setting and Training Details} For the Semantic Extractor and Pixel Enhancer, we maintain the same settings as the released code of the baseline network~\cite{Zou_2023_CVPR, DBLP:conf/nips/HeoHOLK22, DBLP:conf/nips/ShiGXWYD22}. Additionally, to align the semantic-level information with pixel-level features, we introduce a channel compression module, projecting the semantic-level features from 512 to 120. We train our model with five input frames ($T= 5$) sampled from the same video, and set the input patch size of LR frames as $64 \times 64$. For optimization, we use AdamW with $\beta_1 = 0.9, \beta_2 = 0.99$ and weight decay $=10^{-4}$. The learning rate is initialized to $2 \times 10^{-4}$. The Charbonnier loss is applied on whole frames between the ground-truth $I^{HR}$ and restored frame $I^{SR}$, defined as $L= \sqrt{||I^{HR}-I^{SR} ||^2 +\epsilon^2}$. The whole model is trained based on X-Decoder pre-training at the image-level. Firstly, the X-Decoder with frozen weights is extended to the video-level using the video instance segmentation task. Subsequently, fine-tuning is performed on the entire Semantic Lens, where all parameters of the Semantic Extractor are kept fixed and not updated. The model is implemented with PyTorch-2.0 and trained across 4 NVIDIA 3090 GPUs, each accommodating 2 video clips.

\subsection{Performance Comparison}

\begin{figure}[!t]
\centering
\includegraphics[width=0.95\columnwidth]{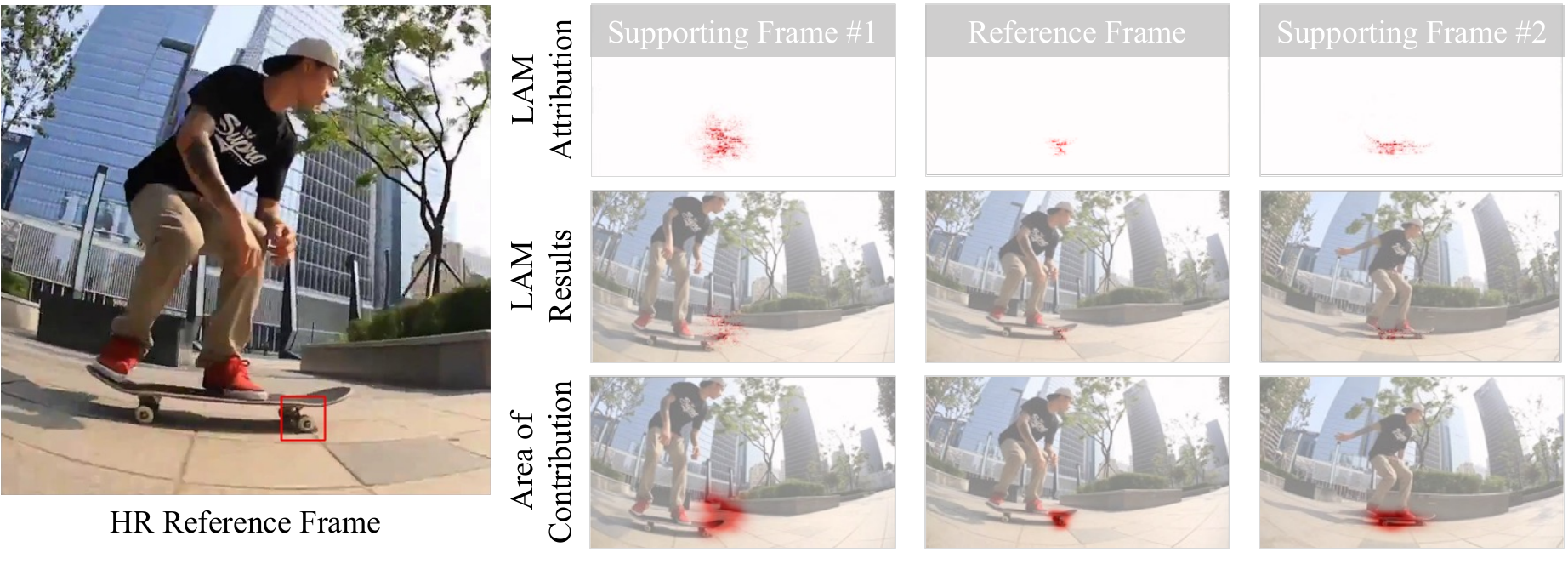}
\caption{The attribution results of adjacent frames.}
\label{fig:lam}
\end{figure}

We compare Semantic Lens with several state-of-the-art methods, including 4 CNN-based VSR methods and 4 transformer-based methods. For all previous methods, we conducted evaluations using their official codes and weights. As shown in Table \ref{tab:experiment}, Semantic Lens achieves state-of-the-art performance on all datasets under both degradations. Specifically, Semantic Lens provides an average improvement of 0.61 dB/0.31 dB/0.27 dB on YTVIS-19/21/22 datasets respectively, establishing the SOTA. What's more, our model achieves better performance in SSIM values.

The qualitative comparisons presented in Figure \ref{fig:ytvis} and Figure \ref{fig:ytvis-22} demonstrate the superiority of the Semantic Lens to restore finer details compared to other methods. While CNN-based methods are still struggling to restore accurate patterns, transformer-based ones, benefiting from their powerful modeling capabilities, have been dedicated to recovering more realistic textures. Compared to PSRT, the textures super-resolved by Semantic Lens is closer to the ground truth, even the distant tiny pattern can be restored intactly. This is achieved by introducing the semantic priors to enable the model to understand the restored contents.

Furthermore, we employ the interpretability tool, Local Attribution Map (LAM)~\cite{DBLP:conf/cvpr/GuD21}, to shed light on the behavior of the Semantic Lens. Given the specified target patch on the reference frame, LAM tracks the information used by the model. It highlights the pixels in multi-frames that contribute to the super-resolution most, generating the corresponding attribution maps. Representative results are shown in Figure \ref{fig:lam}. It can be observed that Semantic Lens, empowered by semantic priors, tends to concentrate more on the same instance across frames, which is particularly evident in the attribution map of supporting frame \#2. 

\begin{table}[!ht]
\small
\centering
\setlength{\tabcolsep}{1.8mm}{
\begin{tabular}{c|cccc}
\hline
& BasicVSR & BasicVSR++ & PSRT & Semantic Lens  \\
\hline
PSNR &36.64 & 36.67 & \underline{37.15} & \textbf{37.19}  \\
SSIM& 0.9495 & 0.9436 & \underline{0.9538} & \textbf{0.9542}  \\
\hline
\end{tabular}}
\caption{Quantitative comparison on the YTVIS-19 dataset.}
\label{tab:finetune}
\end{table}

To further validate the superiority of Semantic Lens, we train BasicVSR, BasicVSR++ and PSRT on the YTVIS dataset, adhering to the training strategy outlined in the original papers, with an input of 5 video frames. As reported in Table \ref{tab:finetune}, we can observe that Semantic Lens achieves consistent and significant performance gain over each other method. It surpasses BasicVSR++ by a margin of up to 0.52 dB in PSNR.

\subsection{Ablation Study}

To verify the effectiveness of each component in the proposed Semantic Lens, we start with a baseline and gradually insert the components. Specifically, we remove the IMAGE and SPACE, retaining only MFSAB in the feature propagation branch as the baseline, while the rest of the network remains the same. From Table \ref{tab:ablation}, it is apparent that each component brings considerable improvement, ranging from 0.05 dB to 0.3 dB in PSNR. Hence, it verifies that beneficial semantic priors for super-resolution are embedded into the pixel-level features.

\begin{table}[!ht]
\small
\centering
\setlength{\tabcolsep}{2.6mm}{
\begin{tabular}{c|cccc|c}
\hline
& Baseline & GPS & ISEE & IMAGE & PSNR/SSIM \\
\hline
1 &\checkmark & & & & 36.75/0.9493  \\
\hline
2&\checkmark & \checkmark & & & 36.84/0.9510\\
3&\checkmark & \checkmark  &  \checkmark & & 37.14/0.9541\\
\hline
4&\checkmark & \checkmark & \checkmark & \checkmark & 37.19/0.9542 \\
\hline
\end{tabular}}
\caption{Results of ablation studies on the YTVIS-19 dataset.}
\label{tab:ablation}
\end{table} 

\begin{figure}[!h]
	\centering
	\includegraphics[width=0.95\columnwidth]{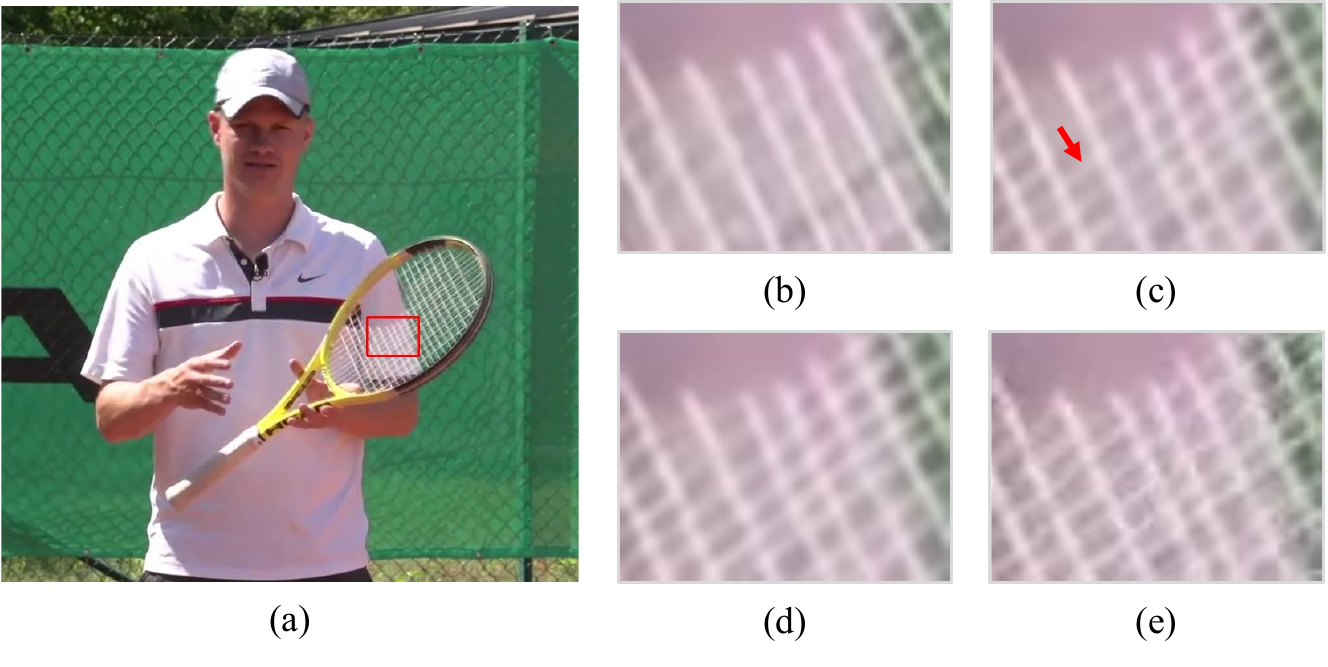}
	\caption{Visual comparisons of ablation for investigating the contribution of key modules. (a) HR Frame. (b) Baseline. (c) Baseline+SPACE. (d) Semantic Lens. (e) Patch of GT.}
	\label{fig:abs}
\end{figure}

As shown in Table \ref{tab:ablation} and Figure \ref{fig:abs}, the proposed SPACE plays an important role in the final super-resolved result because of the adaptive embedding of diverse semantics. Compared with the baseline, the model with SPACE can achieve a more intact and natural recovery on tiny and fine textures (such as strings of the tennis racket in Figure \ref{fig:abs} (c)). It also verifies that introducing IMAGE can indeed help improve the sharpness of the final results of videos.

\section{Conclusion}

In this paper, we introduce a video super-resolution paradigm, named Semantic Lens, that draws knowledge from semantic information. It consists of a Semantic Extractor and a Pixel Enhancer, which are responsible for extracting semantic priors and enhancing pixel-level features from low-resolution videos, respectively. To bridge the semantic and pixel-level information, we deploy a Semantics-Powered Attention Cross-Embedding Block ahead of the feature propagation basic block within the Pixel Enhancer. It iteratively embeds global semantics and instance-specific semantics into the features of VSR, aligning adjacent frames in the instance-centric semantic space.

\section*{Acknowledgements}

This work is supported in part by the National Key Research and Development Program of China under Grant 2022ZD0118001; and in part by the National Natural Science Foundation of China under Grant 62120106009, Grant 62372036, Grant 62332017, and Grant 61972028.

\bibliography{aaai24}

\begin{thebibliography}{29}
\providecommand{\natexlab}[1]{#1}

\bibitem[{Cao et~al.(2022)Cao, Liang, Zhang, Li, Zhang, Wang, and Gool}]{cao2022reference}
Cao, J.; Liang, J.; Zhang, K.; Li, Y.; Zhang, Y.; Wang, W.; and Gool, L.~V. 2022.
\newblock Reference-Based Image Super-resolution with Deformable Attention Transformer.
\newblock In \emph{ECCV}, 325--342.

\bibitem[{Chan et~al.(2021)Chan, Wang, Yu, Dong, and Loy}]{DBLP:conf/cvpr/ChanWYDL21}
Chan, K. C.~K.; Wang, X.; Yu, K.; Dong, C.; and Loy, C.~C. 2021.
\newblock BasicVSR: The Search for Essential Components in Video Super-resolution and Beyond.
\newblock In \emph{CVPR}, 4947--4956.

\bibitem[{Chan et~al.(2022)Chan, Zhou, Xu, and Loy}]{DBLP:conf/cvpr/ChanZXL22a}
Chan, K. C.~K.; Zhou, S.; Xu, X.; and Loy, C.~C. 2022.
\newblock BasicVSR++: Improving Video Super-resolution with Enhanced Propagation and Alignment.
\newblock In \emph{CVPR}, 5962--5971.

\bibitem[{Cheng et~al.(2022)Cheng, Misra, Schwing, Kirillov, and Girdhar}]{DBLP:conf/cvpr/ChengMSKG22}
Cheng, B.; Misra, I.; Schwing, A.~G.; Kirillov, A.; and Girdhar, R. 2022.
\newblock Masked-Attention Mask Transformer for Universal Image Segmentation.
\newblock In \emph{CVPR}, 1280--1289.

\bibitem[{Deudon et~al.(2020)Deudon, Kalaitzis, Goytom, Arefin, Lin, Sankaran, Michalski, Kahou, Cornebise, and Bengio}]{deudon2020highresnet}
Deudon, M.; Kalaitzis, A.; Goytom, I.; Arefin, M.~R.; Lin, Z.; Sankaran, K.; Michalski, V.; Kahou, S.~E.; Cornebise, J.; and Bengio, Y. 2020.
\newblock HighRes-Net: Recursive Fusion for Multi-Frame Super-resolution of Satellite Imagery.
\newblock arXiv:2002.06460.

\bibitem[{Gu and Dong(2021)}]{DBLP:conf/cvpr/GuD21}
Gu, J.; and Dong, C. 2021.
\newblock Interpreting Super-resolution Networks With Local Attribution Maps.
\newblock In \emph{CVPR}, 9199--9208.

\bibitem[{He et~al.(2017)He, Gkioxari, Doll{\'{a}}r, and Girshick}]{DBLP:conf/iccv/HeGDG17}
He, K.; Gkioxari, G.; Doll{\'{a}}r, P.; and Girshick, R.~B. 2017.
\newblock Mask {R-CNN}.
\newblock In \emph{ICCV}, 2980--2988.

\bibitem[{Heo et~al.(2022)Heo, Hwang, Oh, Lee, and Kim}]{DBLP:conf/nips/HeoHOLK22}
Heo, M.; Hwang, S.; Oh, S.~W.; Lee, J.; and Kim, S.~J. 2022.
\newblock {VITA:} Video Instance Segmentation via Object Token Association.
\newblock In \emph{NeurIPS}, 23109--23120.

\bibitem[{Kirillov et~al.(2019)Kirillov, He, Girshick, Rother, and Doll{\'{a}}r}]{DBLP:conf/cvpr/KirillovHGRD19}
Kirillov, A.; He, K.; Girshick, R.~B.; Rother, C.; and Doll{\'{a}}r, P. 2019.
\newblock Panoptic Segmentation.
\newblock In \emph{CVPR}, 9404--9413.

\bibitem[{Li et~al.(2023)Li, Ji, Qin, Niu, Ren, Afghah, Guo, and Ma}]{Li_2023_CVPR}
Li, G.; Ji, J.; Qin, M.; Niu, W.; Ren, B.; Afghah, F.; Guo, L.; and Ma, X. 2023.
\newblock Towards High-Quality and Efficient Video Super-Resolution via Spatial-Temporal Data Overfitting.
\newblock In \emph{CVPR}, 10259--10269.

\bibitem[{Liang et~al.(2022)Liang, Fan, Xiang, Ranjan, Ilg, Green, Cao, Zhang, Timofte, and Gool}]{DBLP:conf/nips/LiangFXRIGC0TG22}
Liang, J.; Fan, Y.; Xiang, X.; Ranjan, R.; Ilg, E.; Green, S.; Cao, J.; Zhang, K.; Timofte, R.; and Gool, L.~V. 2022.
\newblock Recurrent Video Restoration Transformer with Guided Deformable Attention.
\newblock In \emph{NeurIPS}, 378--393.

\bibitem[{Liu et~al.(2022{\natexlab{a}})Liu, Yang, Fu, and Qian}]{DBLP:conf/cvpr/Liu0FQ22}
Liu, C.; Yang, H.; Fu, J.; and Qian, X. 2022{\natexlab{a}}.
\newblock Learning Trajectory-Aware Transformer for Video Super-Resolution.
\newblock In \emph{CVPR}, 5677--5686.

\bibitem[{Liu et~al.(2022{\natexlab{b}})Liu, Yang, Fu, and Qian}]{liu2022ttvfi}
Liu, C.; Yang, H.; Fu, J.; and Qian, X. 2022{\natexlab{b}}.
\newblock TTVFI: Learning Trajectory-Aware Transformer for Video Frame Interpolation.
\newblock arXiv:2207.09048.

\bibitem[{Liu et~al.(2021)Liu, Zhao, Ruan, Shang, and Liu}]{DBLP:conf/aaai/LiuZRS021}
Liu, H.; Zhao, P.; Ruan, Z.; Shang, F.; and Liu, Y. 2021.
\newblock Large Motion Video Super-resolution with Dual Subnet and Multi-Stage Communicated Upsampling.
\newblock In \emph{AAAI}, 2127--2135.

\bibitem[{Liu et~al.(2023{\natexlab{a}})Liu, Jin, Yao, Lin, and Zhao}]{DBLP:journals/tcsv/LiuJYLZ23}
Liu, M.; Jin, S.; Yao, C.; Lin, C.; and Zhao, Y. 2023{\natexlab{a}}.
\newblock Temporal Consistency Learning of Inter-Frames for Video Super-resolution.
\newblock \emph{{IEEE} Transactions on Circuits and Systems for Video Technology}, 33(4): 1507--1520.

\bibitem[{Liu et~al.(2023{\natexlab{b}})Liu, Xu, Yao, Lin, and Zhao}]{10255610}
Liu, M.; Xu, C.; Yao, C.; Lin, C.; and Zhao, Y. 2023{\natexlab{b}}.
\newblock JNMR: Joint Non-linear Motion Regression for Video Frame Interpolation.
\newblock \emph{IEEE Transactions on Image Processing}, 32: 5283--5295.

\bibitem[{Lu et~al.(2023)Lu, Liu, Jin, Luo, Yue, Wang, Zuo, Zeng, Fan, Pang et~al.}]{lu2023virtual}
Lu, Z.; Liu, Y.; Jin, M.; Luo, X.; Yue, H.; Wang, Z.; Zuo, S.; Zeng, Y.; Fan, J.; Pang, Y.; et~al. 2023.
\newblock Virtual-Scanning Light-Field Microscopy for Robust Snapshot High-Resolution Volumetric Imaging.
\newblock \emph{Nature Methods}, 20(5): 735--746.

\bibitem[{Shi et~al.(2022)Shi, Gu, Xie, Wang, Yang, and Dong}]{DBLP:conf/nips/ShiGXWYD22}
Shi, S.; Gu, J.; Xie, L.; Wang, X.; Yang, Y.; and Dong, C. 2022.
\newblock Rethinking Alignment in Video Super-resolution Transformers.
\newblock In \emph{NeurIPS}, 36081--36093.

\bibitem[{Wan et~al.(2022)Wan, Zhang, Chen, and Liao}]{DBLP:conf/cvpr/Wan00022}
Wan, Z.; Zhang, B.; Chen, D.; and Liao, J. 2022.
\newblock Bringing Old Films Back to Life.
\newblock In \emph{CVPR}, 17673--17682.

\bibitem[{Wang et~al.(2019)Wang, Chan, Yu, Dong, and Loy}]{DBLP:conf/cvpr/WangCYDL19}
Wang, X.; Chan, K. C.~K.; Yu, K.; Dong, C.; and Loy, C.~C. 2019.
\newblock {EDVR:} Video Restoration With Enhanced Deformable Convolutional Networks.
\newblock In \emph{CVPRW}, 1954--1963.

\bibitem[{Xu et~al.(2023)Xu, Yu, Wang, Mi, and Yao}]{xu2023implicit}
Xu, K.; Yu, Z.; Wang, X.; Mi, M.~B.; and Yao, A. 2023.
\newblock An Implicit Alignment for Video Super-resolution.
\newblock arXiv:2305.00163.

\bibitem[{Yang et~al.(2020)Yang, Yang, Fu, Lu, and Guo}]{yang2020learning}
Yang, F.; Yang, H.; Fu, J.; Lu, H.; and Guo, B. 2020.
\newblock Learning Texture Transformer Network for Image Super-resolution.
\newblock In \emph{CVPR}, 5791--5800.

\bibitem[{Yang et~al.(2021)Yang, Fan, Fu, and Xu}]{ytvis21dataset}
Yang, L.; Fan, Y.; Fu, Y.; and Xu, N. 2021.
\newblock The 3rd Large-scale Video Object Segmentation Challenge-video instance segmentation track.
\newblock \url{https://youtube-vos.org/dataset/vis/}.

\bibitem[{Yang, Fan, and Xu(2019)}]{DBLP:conf/iccv/YangFX19}
Yang, L.; Fan, Y.; and Xu, N. 2019.
\newblock Video Instance Segmentation.
\newblock In \emph{ICCV}, 5187--5196.

\bibitem[{Yang, Fan, and Xu(2022)}]{ytvis22dataset}
Yang, L.; Fan, Y.; and Xu, N. 2022.
\newblock The 4th Large-scale Video Object Segmentation Challenge-video instance segmentation track.
\newblock \url{https://youtube-vos.org/challenge/2022/}.

\bibitem[{Zhang et~al.(2022)Zhang, Gao, Fang, Jiao, and Wei}]{zhang2022mining}
Zhang, Z.; Gao, G.; Fang, Z.; Jiao, J.; and Wei, Y. 2022.
\newblock Mining Unseen Classes via Regional Objectness: A Simple Baseline for Incremental Segmentation.
\newblock In \emph{NeurIPS}, 24340--24353.

\bibitem[{Zhou et~al.(2022)Zhou, Li, Lu, Han, and Lu}]{DBLP:conf/cvpr/ZhouLL0L22}
Zhou, K.; Li, W.; Lu, L.; Han, X.; and Lu, J. 2022.
\newblock Revisiting Temporal Alignment for Video Restoration.
\newblock In \emph{CVPR}, 6043--6052.

\bibitem[{Zhu et~al.(2023)Zhu, Wei, Liang, Zhang, and Zhao}]{zhu2023ctp}
Zhu, H.; Wei, Y.; Liang, X.; Zhang, C.; and Zhao, Y. 2023.
\newblock CTP: Towards Vision-Language Continual Pretraining via Compatible Momentum Contrast and Topology Preservation.
\newblock In \emph{ICCV}, 22257--22267.

\bibitem[{Zou et~al.(2023)Zou, Dou, Yang, Gan, Li, Li, Dai, Behl, Wang, Yuan, Peng, Wang, Lee, and Gao}]{Zou_2023_CVPR}
Zou, X.; Dou, Z.-Y.; Yang, J.; Gan, Z.; Li, L.; Li, C.; Dai, X.; Behl, H.; Wang, J.; Yuan, L.; Peng, N.; Wang, L.; Lee, Y.~J.; and Gao, J. 2023.
\newblock Generalized Decoding for Pixel, Image, and Language.
\newblock In \emph{CVPR}, 15116--15127.

\end{thebibliography}

\end{document}